\documentstyle[gb4e,psfig,colacl]{article}

\newcommand{\ctxt}[1]{\begin{center} #1 \end{center}} 

\title{A Linguistically Interpreted Corpus of German Newspaper Text}

\author{Wojciech Skut, Thorsten Brants, Brigitte Krenn, Hans Uszkoreit\\
        Universit{\"a}t des Saarlandes, Computational Linguistics\\
        D-66041 Saarbr{\"u}cken, Germany\\
        {\tt \{skut,brants,krenn,uszkoreit\}@coli.uni-sb.de}\\[1ex]
        {\em In ESSLLI-98 Workshop on Recent Advances in Corpus
        Annotation, Saarbr{\"u}cken, 1998}}

\begin{document}

\maketitle

\renewcommand{\thefootnote}{\fnsymbol{footnote}}
\footnotetext[1]{This is a revised version of the paper \cite{SkutEA:97c}.}
\footnotetext[2]{The work has been carried out in the project NEGRA of the Sonderforschungsbereich 378 `Kognitive ressourcenadaptive Prozesse' (resource adaptive cognitive processes) funded by the Deutsche Forschungsgemeinschaft.}
\renewcommand{\thefootnote}{\arabic{footnote}}

\begin{abstract}
\small

In this paper, we report on the development of an annotation scheme and
annotation tools for unrestricted German text. Our representation
format is based on argument structure, but also permits the extraction
of other kinds of representations. We discuss several methodological
issues and the analysis of some phenomena. Additional focus is on the
tools developed in our project and their applications.

\normalsize
\end{abstract}

\section{Introduction}

Parts of a German newspaper corpus, the Frankfurter Rundschau, have
been annotated with syntactic structure. The raw text has been taken
from the multilingual CD-ROM which has been produced by the European
Coding Initiative ECI, and is distributed by the Linguistic Data
Consortium LDC.

The aim is to create a linguistically interpreted text corpus, thus
setting up a basis for corpus linguistic research and statistics-based
approaches for German. We developed tools to facilitate
annotations. These tools are easily adaptable to other annotation
schemes.

\section{Corpora for Data-Driven NLP}

An important pardigm shift is currently taking place in linguistics
and language technology. Purely introspective research focussing on a
limited number of isolated phenomena is being replaced by a more
data-driven view of language. The growing importance of 
stochastic methods opens new avenues for dealing with the wealth of
phenomena found in real texts, especially phenomena requiring a model
of preferences or degrees of grammaticality.


This new research paradigm requires very large corpora annotated with
different kinds of linguistic information. Since the main objective
here is rich, transparent and consistent annotation rather than
putting forward hypotheses or explanatory claims, the following
requirements are often stressed:

\begin{description}

\item[descriptivity:] phenomena should be described rather 
        than explained as explanatory mechanisms can be derived
        (induced) from the data.

\item[data-drivenness:] the formalism should provide means for 
        representing all types of grammatical constructions occurring
        in the corpus%
\footnote{This is what distinguishes corpora used for grammar induction 
from other collections of language data. For instance, so-called {\em
test suites} (cf. \cite{LehmannEA:96}) consist of typical instances of
selected phenomena and thus focus on a subset of real-world language.}.

\item[theory-neutrality:] the annotation format should not be 
        influenced by theory-internal considerations. However,
        annotations should contain enough information to permit the
        extraction of theory-specific representations.

\end{description}

In addition, the architecture of the annotation scheme should make it
easy to refine the information encoded, both in width (adding new
description levels) and depth (refining existing
representations). Thus a structured, multi-stratal organisation of the
representation formalism is desirable.

The representations themselves have to be easy to determine on the
basis of simple empirical tests, which is crucial for the consistency
and a reasonable speed of annotation.

\section{Why Tectogrammatical Structure?}

In the data-driven approach, the choice of a particular representation
formalism is an engineering problem rather than a matter of
`adequacy'. More important is the theory-independence and reusability
of linguistic knowledge, i.e., the recoverability of
theory/application specific representations, which in the area of NL
syntax fall into two classes:

\begin{description}

\item[Phenogrammatical structure:] the structure reflecting surface order, 
e.g. {\em constituent structure} or topological models of surface
syntax, cf. \cite{Ahrenberg:agcpsafs}, \cite{Reape:duawovig}.

\item[Tectogrammatical representations:] predicate-argument structures
reflecting lexical argument structure and providing a guide for
assembling meanings. This level is present in almost every theory:
D-structure (GB), f-structure (LFG) or argument structure (HPSG). A
theory based mainly on tectogrammatical notions is dependency grammar,
cf. \cite{Tesniere:edss}.

\end{description}
As annotating both structures separately presents substantial effort,
it is better to recover constituent structure automatically from an
argument structure treebank, or vice versa. Both alternatives are
discussed in the following sections.

\subsection{Annotating Constituent Structure}

Phenogrammatical annotations require an additional mechanism encoding
tectogrammatical structure, e.g., trace-filler dependencies
representing discontinuous constituents in a context-free constituent
structure (cf. \cite{Marcusea:94}, \cite{Sampson95}). A major drawback
for annotation is that such a hybrid formalism renders the structure
less transparent, as is the phrase-structure representation of
sentence (\ref{weinen}):

\small
\begin{exe}
\ex \gll daran wird ihn Anna erkennen, dass er weint \\
        at-it will him Anna recognise that he cries\\
        `Anna will recognise him at his cry' \label{weinen}
\end{exe}
\normalsize

\psfig{file=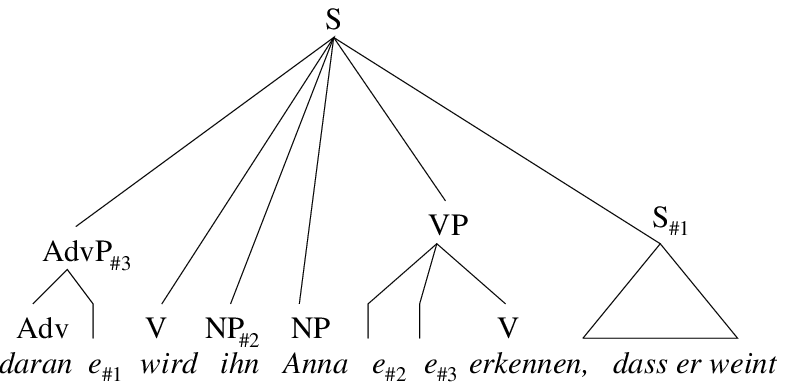,width=6.8cm}

Furthermore, the descriptivity requirement could be difficult to meet
since constituency has been used as an explanatory device for several
phenomena (binding, quantifier scope, focus projection).

The above remarks carry over to other models of {\em phenogrammatical
structure}, e.g. {\em topological fields},
cf. \cite{Bech:suddvi}. A sample structure is given below%
\footnote{LSB, RSB stand for {\em left} and {\em right sentence bracket}.}

\medskip

\psfig{file=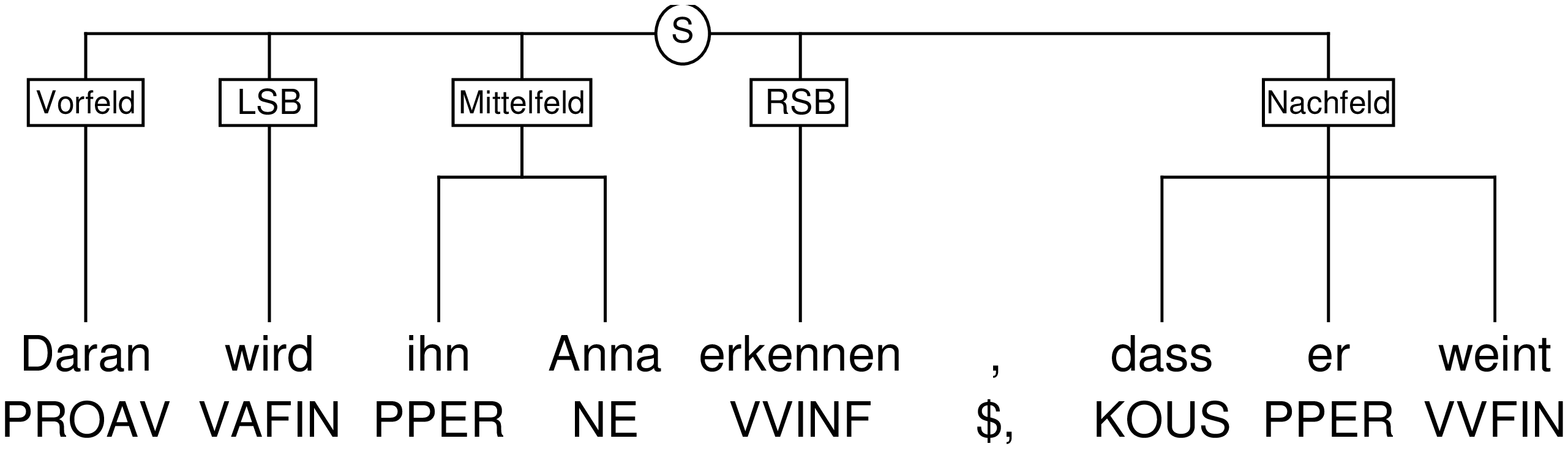,width=8.2cm,angle=-90}

Here, as well, topological information is insufficient to express the
underlying tectogrammatical structure (e.g., the attachment of the
extraposed that-clause)%
\footnote{Even annotating grammatical functions is not enough as long 
as we do not explicitly encode their tectogrammatical attachment of
such functions.}. Thus the {\em field model} can be viewed as a
non-standard phrase-structure grammar which needs additional
tectogrammatical annotations.

\subsection{Argument Structure Annotations}

An alternative to annotating surface structure is to directly specify
the tectogrammatical structure, as shown in the following figure:

\psfig{file=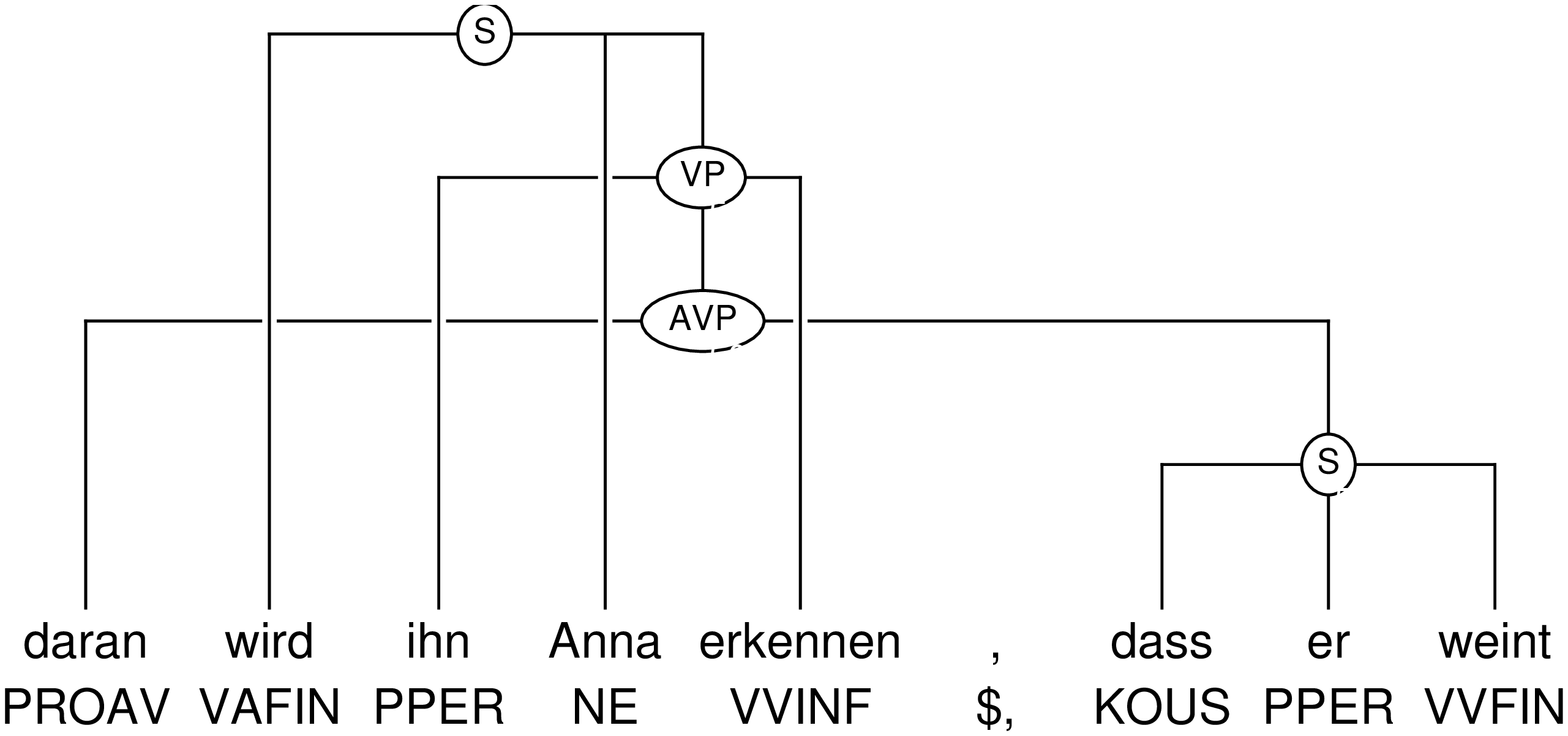,width=8.2cm,angle=-90}

This encoding has several advantages. Local and non-local dependencies
are represented in a uniform way. Discontinuity does not influence the
hierarchical structure, so the latter can be determined on the basis
of lexical subcategorisation requirements, agreement and some semantic
information.

An important advantage of tectogrammatical structure is its proximity
to semantics. This kind of representations is also more theory-neutral
since most differences between syntactic theories occur at the
phenogrammatical level, the tectogrammatical structures being fairly
similar.

Furthermore, a constituent tree can be recovered from a
tectogrammatical structure. Thus tectogrammatical representations
provide a uniform encoding of information for which otherwise both
constituent trees {\em and} trace-filler annotations are needed.

Apart from the work reported in this paper, tectogrammatical
annotations have been successfully used in the TSNLP project to
construct a language competence database, cf. \cite{LehmannEA:96}.

\subsection{Suitability for German}

Further advantages of tectogrammatical annotations have to do with the
fairly weak constraints on German word order, resulting in a good deal
of discontinuous constituency. This feature makes it difficult to come
up with a precise notion of constituent structure. In the effect,
different kinds of structures are proposed for German, the criteria
being often theory-internal%
\footnote{Flat or binary right-recursive structures, not to mention 
the status of the head in verb-initial, verb-second and verb-final
clauses, cf. \cite{Netter:onhnm}, \cite{Kasper:aitm},
\cite{Nerbonne:94}, \cite{Pollard:96}.}. 

In addition, phrase-structure annotations augmented with the many
trace-filler co-references would lack the transparency desirable for
ensuring the consistency of annotation.

\section{Methodology}

The standard methodology of determining constituent structure (e.g.,
the {\em Vorfeld} test) does not carry over to tectogrammatical
representations, at least not in all its aspects. The following
sections are thus concerned with methodological issues.

\subsection{Structures vs. Labels}

The first question to be answered here is how much information has to
be encoded structurally. Rich structures usually introduce high
spurious ambiguity potential, while flat representations (e.g.,
category or function labels) are significantly easier to manipulate
(alteration, refinement, etc.).

Thus it is a good strategy to use rather simple structures and
express more information by labels.

\subsection{Structural Representations}
\label{str1}

As already mentioned, tectogrammatical structures are often thought of
in terms of {\em dependency grammar} (DG, cf. \cite{Hudson:wg},
\cite{Hellwig:cpattsafp}), which might suggest using conventional 
{\em dependency trees} (stemmas) as our representation format.
However, this would impose a number of restrictions that follow from
the theoretical assumptions of DG. It is mainly the DG notion of heads
that creates problems for a flexible and maximally theory-neutral
approach. In a conventional dependency tree, heads have to be unique,
present and of lexical status, requirements other theories might not
agree with.

That is why we prefer a representation format in which heads are
distinguished outside the structural component, as shown in the figure
below, sentence (\ref{baecker})%
\footnote{Edge labels: HD head, SB subject, OC clausal complement, PD
predicative, MO modifier. Note that crossing edges indicate
discontinuous constituency.}:

\small
\begin{exe}
\ex \gll B\"acker wollte er nie werden\\
        baker wanted he never become\\
        `he never wanted to become a baker' \label{baecker}
\end{exe}
\normalsize

\ctxt{
\mbox{\psfig{file=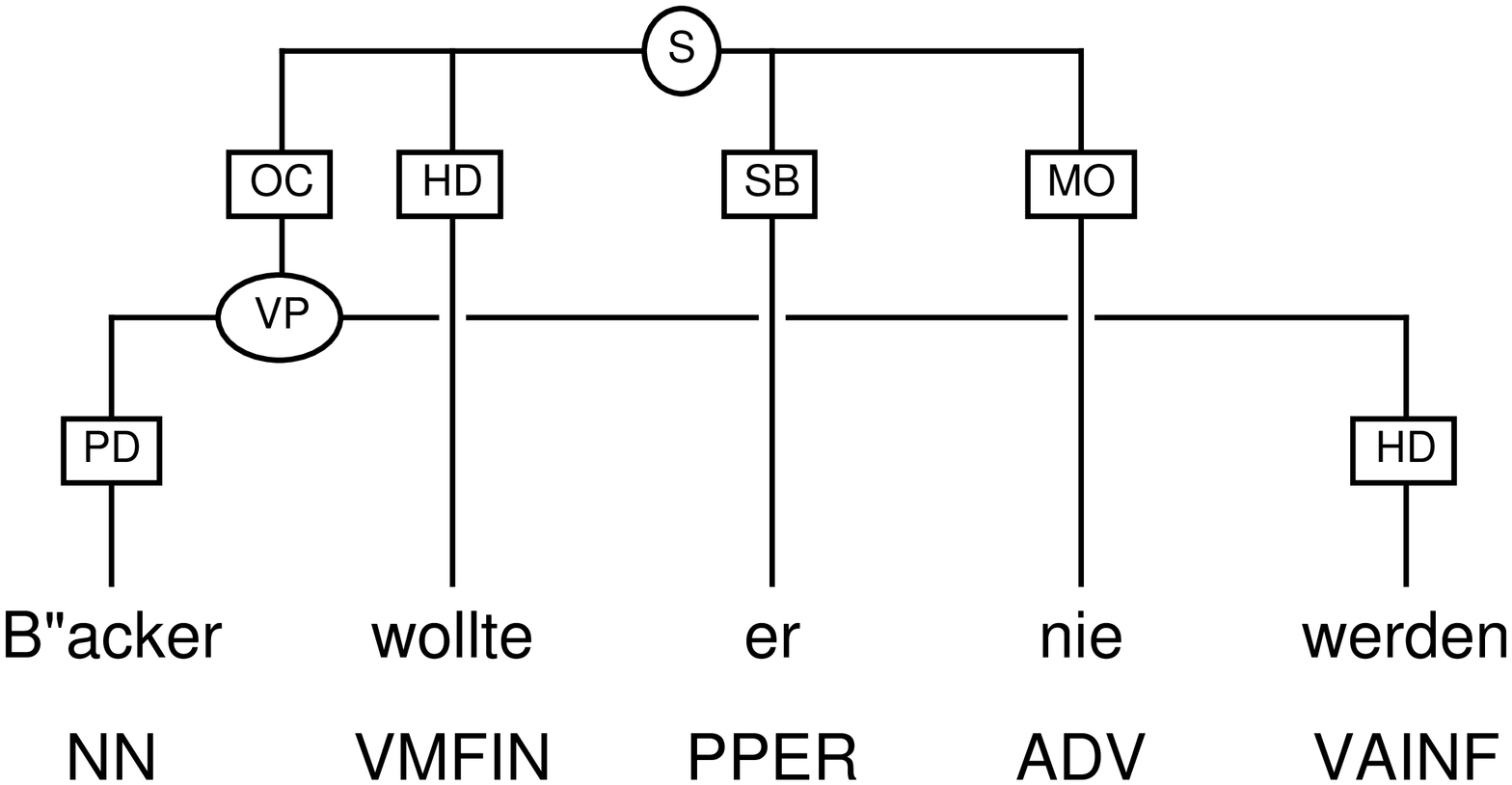,width=7cm,angle=-90}}
} 
The tree encodes three kinds of information:

\begin{description}
\item[tectogrammatical structure:] trees with possibly crossing
branches (no non-tangling condition);

\item[syntactic category:] node labels and part-of-speech tags
(Stuttgart-T{\"u}bingen Tagset, cf. \cite{Schiller;Thielen:94}).

\item[functional annotations:] edge labels.
\end{description}


\subsection{Classification of Labels}

Compared to the fairly simple structures employed by our annotation
scheme, the functional annotations encode a great deal of linguistic
information. We have already stressed that the notion {\em head} is
distinguished at this level. Accordingly, it seems to be the
appropriate stratum to encode the differences between different classes
of dependencies.

For instance, most linguistic theories distinguish between complements
and adjuncts. Unfortunately, the theories do not agree on the criteria
for drawing the line between the two classes of dependents. To this
date there is no single combination of criteria such as category,
morphological marking, optionality, uniqueness of role filling,
thematic role or semantic properties that can be turned into a
transparent operational distinction linguists of different schools
would subscribe to.

In our scheme, we try to stay away from a theoretical commitment
concerning borderline decisions. The distinction between functional
labels such as SB and DA -- standing for traditional grammatical
functions -- on the one hand and phrases labelled MO on the other should
not be interpreted as a classification into complements and adjuncts.
For the time being, functional labels different from MO are assigned
only if the grammatical function of the phrase can easily be detected
on the basis of the linguistic data. MO is used, e.g., to label
adjuncts as well as prepositional objects. Likewise the label OC is
used for easily recognisable clausal complements. Other embedded
sentences depending on the verb are labelled as MO%
\footnote{MO is inspired by the usage of the term
`modifier' in traditional structuralist linguistics where some authors
\cite{Bloomfield:l} use it for adjuncts and others also for complements
\cite{Trubetzkoy:lreldldeld}.}.  This is consistent with our philosophy of
stepwise refinement. We are in the process of designing a more
fine-grained classification of functional labels together with
testable criteria for assigning them. This classification will not
contain a distinction between complements and adjuncts. Thus the
locative phrase {\em in Berlin} in the sentence {\em Peter wohnt in
Berlin} (Peter lives in Berlin) will just be marked as a locative MO
with the category PP. As linguistic theories disagree on the question,
we will not ask the annotators to decide whether this phrase is a
complement of the verb.

This strategy differs from the one pursued by the creators of the Penn
Treebank. There the difference between complements and adjuncts is
encoded in the hierarchical structure. Verbal complements are encoded
as siblings of the verb whereas adjuncts are adjoined at a higher
level. In a case of doubt, the annotators are asked to select
adjunction. We consider this structural encoding less suitable for
refinement than a hierarchy of functional labels in which MO can be
further specified by sublabels.

\section{Annotation Tools}
\label{sec:tool}


The development of linguistically interpreted corpora presents a
laborious and time-consuming task. In order to make the annotation
process more efficient, extra effort has been put into the development of
the annotation software.

\subsection{Structural Annotation}

\begin{figure*}[t]
\begin{center}
\hrule
\bigskip
\centerline{\psfig{file=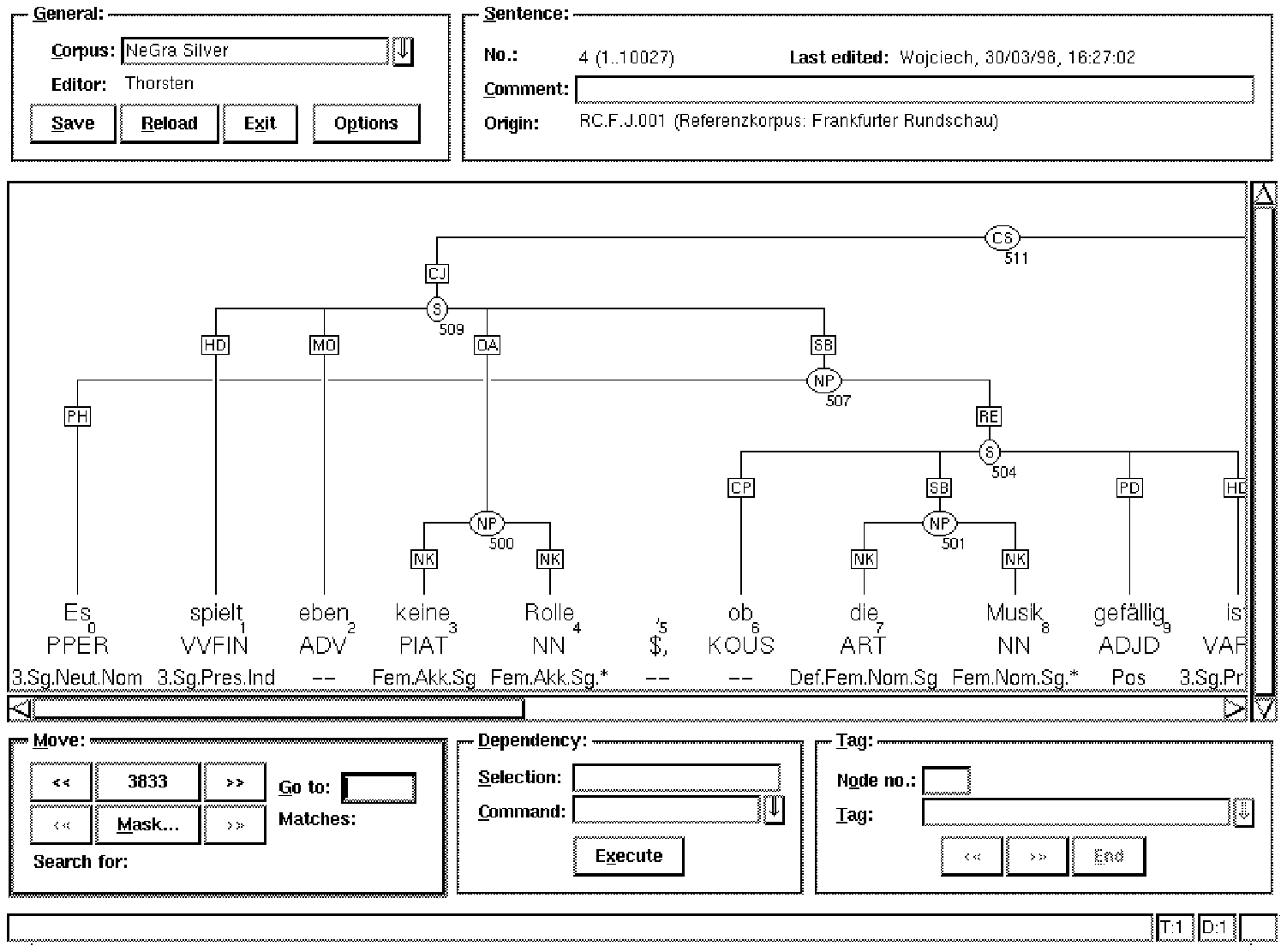,width=15cm}}
\bigskip
\hrule
\mbox{}
\caption{Screen dump of the annotation tool}
\label{FigScreen}
\end{center}
\end{figure*}

The annotation tools are an integrated software package that
communicates with the user via a comfortable graphical interface
\cite{Plaehn:98a}. Both keyboard and mouse input are supported, the structure being
annotated is shown on the screen as a tree. The tools can be employed
for the annotation of different kinds of structures, ranging from our
rudimentary predicate-argument trees to standard phrase structure
annotations with trace-filler dependecies, cf. \cite{Marcusea:94}.  A
screen dump of the annotation tool is shown in figure \ref{FigScreen}.

The kernel part of the annotation tool supports purely manual
annotation. Further modules permit interaction with an external
stochastic or symbolic parser. Thus, the tools are not dependent on a
particular automation method. Also the degree of automation can vary
from part-of-speech tagging and recognition of grammatical functions
to full parsing.

In our project, we rely on an interactive annotation mode in which the
annotator specifies rather small annotation increments that are then
processed by a stochastic parser. The output of the parser is
immediately displayed and the annotator edits it if
necessary. Currently, the annotator's task is to specify substructures
containing up to $20-30$ words; their internal structure as well as
the labels for grammatical functions and categories are assigned by
the parser. The precision of the parser is about $96\%$ for the
assignment of labels and $90\%$ for partial structures
\cite{Brants:Skut:98,Skut:Brants:98a,Skut:Brants:98b}.

Another part of our software package is the corpus search tool. It is
very helpful for both linguistic investigations and detecting
annotation errors. As for this latter application, we have also
developed programs that compare annotations. Each sentence is
annotated independently by two annotators. During the comparison,
inconsistencies are highlighted, and the annotators have to correct
errors and/or agree on one reading.

In addition to the treebank project, the tools are currently used in
the Verbmobil project to annotate transliterated spoken dialogues in
English and German \cite{Stegmann:Hinrichs:98}, in the FLAG project to
annotate spelling errors in German newsgroup texts, and it is planned
to employ them in the DiET project to build a linguistic competence
database \cite{NetterEA:98}.

\subsection{Automation}

The graphical surface communicates with several separate programs to
perform the task of semi-automatic annotation. Currently, these
separate programs are a part-of-speech tagger, a tagger for
grammatical functions and phrasal categories and an {\sf NP}/{\sf PP}
chunker.

The part-of-speech tagger is a trigram part-of-speech tagger that is
trainable for a wide variety of languages and tagsets \cite{Brants:96a}.
We trained it on all previously annotated material in our corpus, using
the Stuttgart-Tübingen tagset, and it currently achieves an accuracy of
96\% on new, unseen text.

In our project, annotation is an interactive task. After the annotator
has specified a partial structure, the tool automatically inserts all
the labels into the structure, i.e. the grammatical functions (edge
labels) and phrasal categories (node labels). This task is performed by
a tagger for grammatical functions and phrasal categories
\cite{Brants:ea:97}. The underlying mechanism is very similar to part
of speech tagging. There, states of a Markov model represent tags, and
outputs represent words. For tagging grammatical functions, states
represent grammatical functions, and outputs represent terminal and
non-terminal tags. Thus, tagging is applied to the next higher level.

Grammatical functions have a different distribution within each type
of phrase, so each type of phrase is modeled by a different Markov
model. If the type of phrase is known, the corresponding model is used
to assign grammatical functions. If the type is not known, all models
run in parallel and the model assigning the highest probability is
used. This determines at the same time the phrasal category. The
tagger is also trained on all previous material of the corpus and
achieves 97\% accuracy for assigning phrasal categories, and 96\%
accuracy for assigning grammatical functions.

When tagging for part-of-speech, grammatical functions, and phrasal
categories, we additionally calculate the second best assignment and
its probability. This is used to estimate the reliability of the first
assignment. If the probability of the alternative is close to that of
the best assignment, the first choice is regarded as unreliable, wheres
it is reliable if the alternative has a much lower
probability. Reliable and unreliable are distinguished by a threshold
on the distance of the best and second best assingment. The annotation
tool simply inserts all reliable labels and asks the human annotator
for confirmation in the unreliable cases.

\begin{figure*}[t]
\begin{center}
\hrule
\bigskip
\centerline{\psfig{file=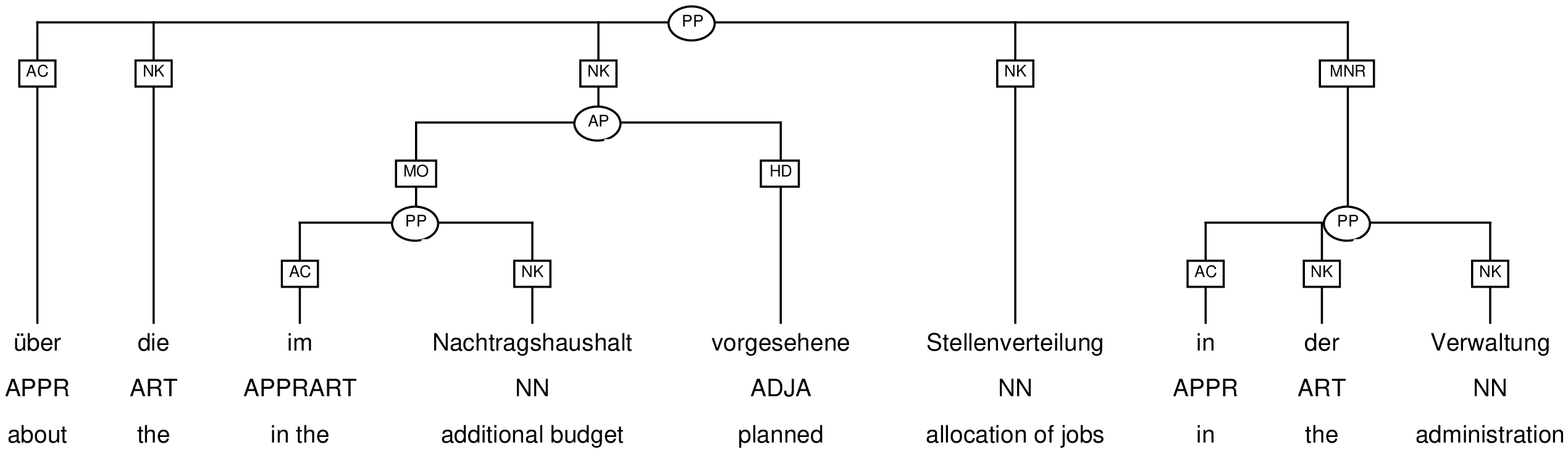,width=\textwidth,angle=-90}}
\bigskip
\hrule
\mbox{}
\caption{Example of a complex {\sf NP}.}
\label{fig:ex-np}
\end{center}
\end{figure*}

The next level of automation is concerned with the structure of {\sf
  NP}s and {\sf PP}s which can be fairly complex in German (see figure
\ref{fig:ex-np}). As shown in \cite{Brants:Skut:98}, recognition of
complete {\sf NP}/{\sf PP} structures can also efficiently performed
with Markov models, encoding {\em relative structures}, i.e. stating
that a word is attached lower, higher or at the same level as its
predecessor. The annotator no longer has to build the structure level
by level, but marks the boundaries of {\sf NP}s and {\sf PP}s, and the
internal structures is generated automatically. This approach has an
accuracy of 85 -- 90\%, depending on the exact task.

\section{Applications of the Corpus}

The corpus provides training and test material for stochastic
approaches to natural language processing. It is also a valuable
source of data for theoretical linguistic investigations, especially
into the relation of competence grammar and language usage.

\subsection{Statistical NLP}

As described in section \ref{sec:tool}, statistical annotation methods
have been developed and implemented. In our bootstrapping approach,
the accuracy of the models is improved and functionality increases as
the annotated corpus grows, thus leading to completely automatic NLP
methods. For instance, the {\em chunk tagger} initially designed to
support the annotator is used for the recognition of major phrases in
unrestricted text pre-tagged with part-of-speech information
\cite{Skut:Brants:98a,Skut:Brants:98b}.

Apart from these applications, the corpus is already used in other
projects to train rule-based and statistical taggers and parsers.

\subsection{Corpus Linguistic Investigations}

The treebank has been successfully used for corpus-linguistic
investigations.  In this regard, two major classes of applications
have arisen so far.  Firstly, a search program enables the user to
find examples of interesting linguistic constructions, which is
especially useful for testing predictions made by linguistic theories.
It has also proved to be a great help in teaching linguistics.

The second, more ambitious class of applications consists in statistical
evaluation of the corpus data.  In a study on relative clause
extrapostion in German \cite{Uszkoreit:ea:98}, we were able to
verify the predictions made by the performance theory of language
formulated by~\newcite{Hawkins:94}.  The corpus data made it possible
to measure the influence of the factors {\em heaviness} and {\em
distance} on the extraposition of relative clauses. The results of
these investigations are also supported by psycholinguistic
experiments.

For investigations on statistics-based collocation extraction, various
portions of the Frankfurter Rundschau Corpus have been automatically
annotated with parts-of-speech and phrase chunks like NP, PP, AP. The
part-of-speech tagger \cite{Brants:96a} and the chunker
\cite{Skut:Brants:98b} have been trained on the annotated and
hand-corrected corpus. 
Although error rates of 10 to 15~\% occur at the stage of chunking, collocation extraction
benefits from structurally annotated corpora because of the
accessibility of syntactic information (1) accuracy of frequency
counts increases, i.e. more syntactically plausible collocation
candidates are found, and (2) grammatical restrictions on collocations
can mostly be automatically derived from the corpus,
cf. \cite{Krenn:98}. 

Syntactically preprocessed corpora are also
a valuable source for insights into actual realisations of
collocations. This is particularly important in the case of partially
flexible collocations. In order to provide material for investigations
into collocations as on the one hand grammatically flexible and on the
other hand lexically fixed constructions, collocation examples found
in syntactically annotated corpora are stored in a database together
with competence-based analyses, cf. \cite{Krenn:98b}.

\section{Conclusions}

The increasing importance of data-oriented NLP requires the development
of a specific methodology, partly different from the generative
paradigm which has dominated linguistics for nearly 40 years. The
importance of consistent and efficient encoding of linguistic
knowledge has absolute priority in this new approach, and thus we have
argued for easing the burden of explanatory claims, which has proved
to be a severe constraint on linguistic formalism.

We have presented a number of linguistic analyses used in our treebank
and examples of the interaction of different syntactic phenomena. We
also have shown how the particular representation chosen enables the
derivation of other, theory specific representations. Finally we have
given examples for applications of the corpus in statistics-based NLP
and corpus linguistics. Our claims are backed by an annotated corpus
of currently about 12,000 sentences, all of which have been annotated
twice in order to ensure consistency.

\end{document}